\documentclass[acmtog]{acmart}

\usepackage{booktabs} %
\usepackage{comment}
\usepackage{tabularx}
\usepackage{multirow}
\usepackage{caption}
\usepackage{subcaption}

\newcommand{\zexp}{\mathbf{z}_{\textrm{exp}}}
\newcommand{\zid}{\mathbf{z_{\textrm{id}}}}

\newcommand{\B}{\mathcal{B}}
\newcommand{\F}{\mathcal{F}}
\newcommand{\G}{\mathcal{G}}

\newcommand{\OURS}{NPGA}

\usepackage[capitalise]{cleveref}
\usepackage[ruled]{algorithm2e} %

\SetAlFnt{\small}
\SetAlCapFnt{\small}
\SetAlCapNameFnt{\small}
\SetAlCapHSkip{0pt}

\acmJournal{TOG}

\copyrightyear{2024}
\acmYear{2024}
\setcopyright{rightsretained}
\acmConference[SA Conference Papers '24]{SIGGRAPH Asia 2024 Conference Papers}{December 3--6, 2024}{Tokyo, Japan}
\acmBooktitle{SIGGRAPH Asia 2024 Conference Papers (SA Conference Papers '24), December 3--6, 2024, Tokyo, Japan}\acmDOI{10.1145/3680528.3687689}
\acmISBN{979-8-4007-1131-2/24/12}

\begin{document}
\title{NPGA: Neural Parametric Gaussian Avatars}

\author{~ Simon Giebenhain}
\affiliation{%
  \institution{Technical University of Munich}
  \country{Germany}
}
\author{~ Tobias Kirschstein}
\affiliation{%
  \institution{Technical University of Munich}
  \country{Germany}
}
\author{~ Martin Rünz}
\affiliation{%
  \institution{Synthesia}
  \country{Germany}
}\author{~ Lourdes Agapito}
\affiliation{%
  \institution{University College London}
  \country{United Kingdom}
}
\author{~ Matthias Nießner}
\affiliation{%
  \institution{Technical University of Munich}
  \country{Germany}
}




\begin{teaserfigure}
    \vspace{-0.3cm}
  \includegraphics[width=\textwidth]{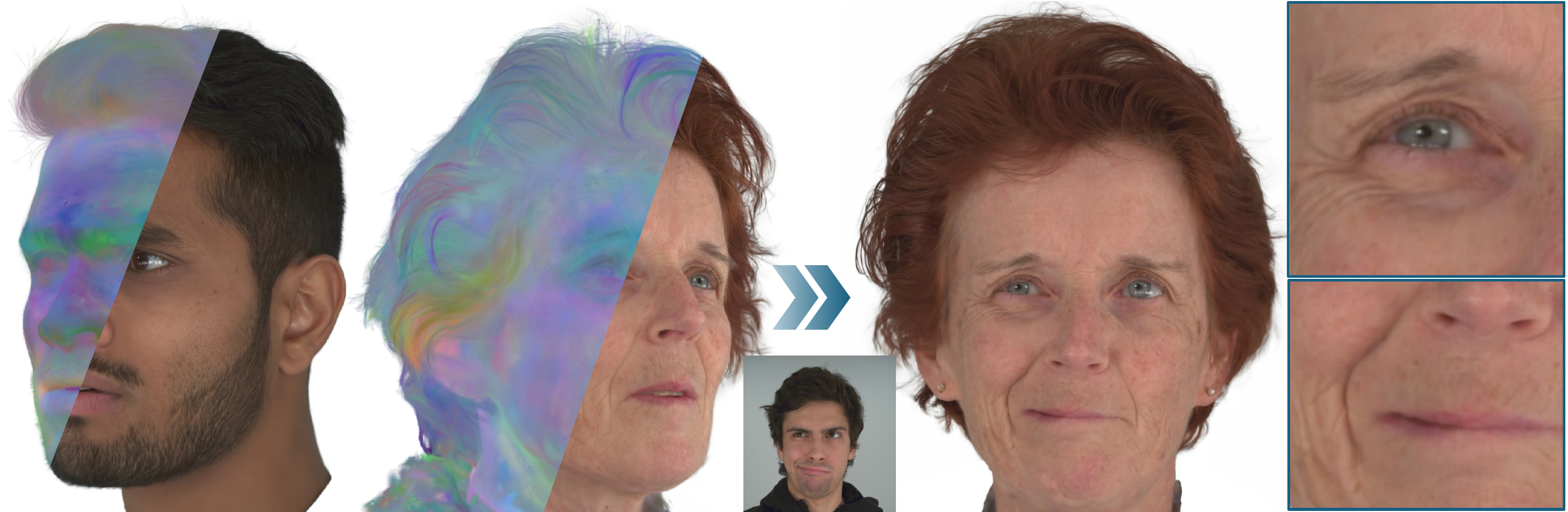}
  \caption{
    \textbf{NPGA:} 
    We utilize the rich expression space of Neural Parametric Head Models to create high-fidelity avatars with fine-grained expression control.
    Our avatars consist of a dynamics module and a canonical Gaussian point cloud, which is augmented with per-primitive features that encode valuable semantic information, as indicated on the left.
    On the right, we demonstrate a highly detailed cross-reenactment using the inset image as a driving expression.}
  \label{fig:teaser}
\end{teaserfigure}
 \begin{abstract}

  The creation of high-fidelity, digital versions of human heads is an important stepping stone in the process of further integrating virtual components into our everyday lives. 
  Constructing such avatars is a challenging research problem, due to a high demand for photo-realism and real-time rendering performance.
  In this work, we propose Neural Parametric Gaussian Avatars (NPGA), a data-driven approach to create high-fidelity, controllable avatars from multi-view video recordings.
  We build our method around 3D Gaussian splatting for its highly efficient rendering and to inherit the topological flexibility of point clouds.
  In contrast to previous work, we condition our avatars' dynamics on the rich expression space of neural parametric head models (NPHM), instead of mesh-based 3DMMs.
  To this end, we distill the backward deformation field of our underlying NPHM into forward deformations which are compatible with rasterization-based rendering. All remaining fine-scale, expression-dependent details are learned from the multi-view videos.
  For increased representational capacity of our avatars, we propose per-Gaussian latent features that condition each primitives dynamic behavior.
  To regularize this increased dynamic expressivity, we propose Laplacian terms on the latent features and predicted dynamics.
  We evaluate our method on the public NeRSemble dataset, demonstrating that NPGA significantly outperforms the previous state-of-the-art avatars on the self-reenactment task by $\approx2.6$~PSNR. Furthermore, we demonstrate accurate animation capabilities from real-world monocular videos.
\end{abstract}

\maketitle

\section{Introduction}

Creating 
{\let\thefootnote\relax\footnotetext{\scriptsize{Project Website: \url{https://simongiebenhain.github.io/NPGA/}}}}
photo-realistic 3D avatars is one of the core challenges in computer graphics and includes a wide range of applications such as movies, games, AR/VR teleconferencing, and the metaverse.
In particular, there is a strong motivation to reconstruct digital avatars from real-world captures, such as multi-view recordings, to obtain a digital copy of a specific real person.
The resulting digital avatars can then be animated and rendered from arbitrary viewpoints while expecting high visual fidelity; e.g., with respect to photo-realistic colors and details, preservation of identity, and the adoption of person-specific mannerisms.
At the same time, many avatar applications demand real-time rendering capabilities without compromising visual quality.

Recent advances at the intersection of computer graphics and vision research have steadily improved methods to digitally reconstruct 3D objects with photo-realistic rendering quality~\cite{mildenhall2021nerf,mueller2022instant,3dgs}. 
In particular, 3D Gaussian Splatting (3DGS)~\cite{3dgs} has been quickly adopted in recent work on digital humans, e.g.~\cite{Zielonka2023Drivable3D,li2024animatablegaussians} and virtual head avatars, e.g.,~\cite{GA,GHA}, due to its efficient rendering and photo-realistic reconstructions. At the same time recent publicly available multi-view datasets~\cite{nersemble,icsik2023humanrf,pan2024renderme,wuu2022multiface}, offer an ever more exciting basis for avatar research.
A central question is how controllability can be achieved.
The most prominent approach for heads is to utilize a 3D morphable model (3DMM), which offers compact descriptions of faces using disentangled parametric spaces for identity and facial expressions.
When utilizing the expressions of an underlying 3DMM, e.g., 
\cite{gafni2021nerface, nha,zielonka2022instant,GA,GHA}, the avatar is optimized to follow a generalized expression space which enables expression transfer or animation through tracking in monocular videos \cite{thies2016face}.
While 3DMMs offer a compact parameterization, their linear nature inherently limits the fidelity of represented expressions.
At the same time, we argue that the underlying expression space plays a crucial role in determining the quality of the created avatars. 
It not only influences the controllability of the resulting avatars but it also limits the sharpness of details. If the stream of input expression codes is insufficiently correlated with the observed images, the optimization problem can become fundamentally ill-posed and lead to overfitting.

To this end, we propose \OURS, a new avatar representation that leverages a learned deformation representation while ensuring that the predicted facial dynamics stay close to the prior of an underlying neural parametric head model (NPHM) \cite{nphm,mononphm}.
NPHM provides our avatars with more fine-grained expression control compared to classical, public 3DMMs~\cite{paysan20093d,FLAME}, which were previously used for avatar creation.
In another line of work \citet{auth_ava} explore a learned neural prior over faces, that helps to map baked uv-space information from a monocular video into a volumetric avatar representation.

NPGA consists of a canonical Gaussian point cloud, that can be forward-deformed using an expression code and rendered using 3DGS, similar to previous work \cite{Zielonka2023Drivable3D,GA,GHA}.
As a rasterization-based approach, 3DGS cannot be efficiently combined with the \emph{backward} deformation field of MonoNPHM. Therefore, we propose a distillation strategy to invert the deformation direction of MonoNPHM's expression prior using a cycle consistency loss, 
similar to SCANimate~\cite{scanimate}.
The resulting \emph{forward} deformation field becomes compatible with the rasterization-based rendering of 3DGS.
To increase the overall dynamic expressivity of our avatars, we further propose to augment our canonical Gaussian with per-Gaussian latent features 
similar to GaussianHeadAvatar~\cite{GHA} and HeadGAS~\cite{dhamo2023headgasrealtimeanimatablehead}.
Compared to these works, we allow these features to influence the movement of Gaussians instead of only influencing dynamic changes in appearance.
This enables our deformation module to operate in a higher dimensional space, that can describe facial dynamics more effectively.
We show that this added expressivity results in higher-fidelity image synthesis, but is required to be appropriately regularized to achieve artifact-free renderings. 
To this end, we formulate Laplacian smoothness terms on the latent features and predicted dynamics, based on the k-nearest neighbor graph in canonical space.
Furthermore, we modify the adaptive density control (ADC) strategy of 3DGS for more detailed avatar reconstructions.

To summarize, our contributions are the following:
\begin{itemize}
    \item To leverage MonoNPHM's rich expression prior for fine-grained animation control and effective optimization, we utilize a cycle-consistency-based distillation strategy.
    \item 
    We condition our deformation network on per-Gaussian latent features for increased dynamic capacity, use Laplacian regularization, and adjust the ADC for a dynamic setting.
    \item We outperform the previous state-of-the-art by 2.6 PSNR and 0.021 SSIM. Furthermore, we demonstrate accurate avatar re-animation from monocular RGB sequences. 
\end{itemize}

\begin{figure*}[htb]

    \centering
    \includegraphics[width=\textwidth]{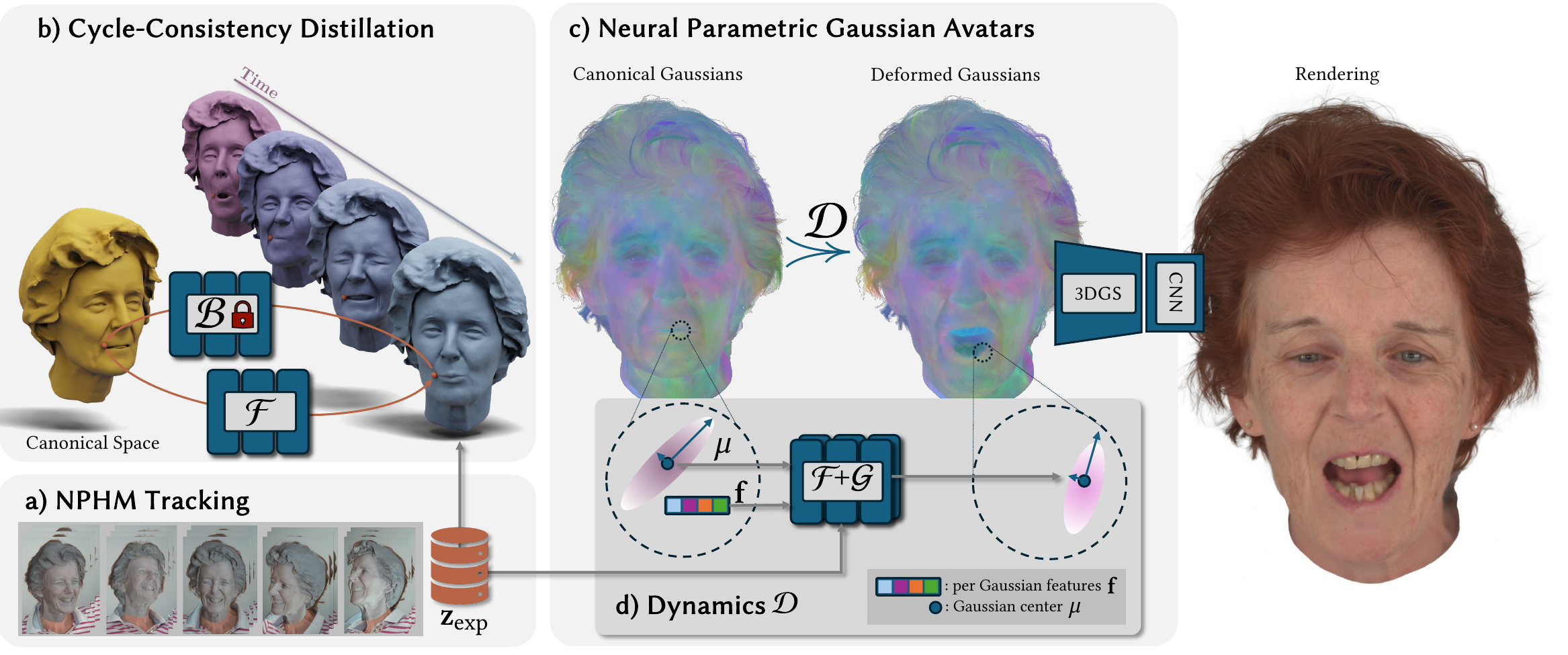}
    \caption{
    \textbf{Method Overview:} 
    The basis of our avatar optimization are multi-view video recordings alongside a MonoNPHM tracking thereof, see (a).
    Next, we extract a forward-deformation prior $\F$ from MonoNPHM's backward deformation field $\B$ using a cycle-consistency loss, see (b). 
    Our avatars consist of a canonical Gaussian point cloud (c), which is warped into posed space using our dynamics module $\mathcal{D}$, consisting of the coarse pre-trained component $\F$ and a detail network $\G$. We condition both networks on per Gaussian features, which dictate each primitive's behavior.
    After rendering the avatar with 3DGS, we employ a screen-space CNN to suppress small-scale artifacts. 
    }
    \label{fig:method_overview}
\end{figure*}

\section{Related Work}

\subsection{Dynamic Scene Representations}

Since humans are inherently dynamic, and subject to topological variation when for example opening and closing the mouth, general dynamic scene representations aim to solve a set of similar problems with the exemption of controllability.
Implicit approaches based on Neural Radiance Fields~\cite{mildenhall2021nerf} provide such flexibility and can be extended to deformable scenes. This extension is either achieved by modeling temporal changes via a deformation field that accompanies a canonical frame~\cite{park2021nerfies,hypernerf}, by adding time as a conditioning variable~\cite{dynerf}, or directly decomposing the 4D scene volume into a computationally manageable representation~\cite{hyperreel,nerfplayer}.
The extensive use of MLPs in neural radiance fields entails a high computational burden, however, and still proves costly after improvements such as hash encodings~\cite{mueller2022instant}, triplanes~\cite{Chan2022_eg3d} or other low-rank approximations~\cite{shao2023tensor4d}.
MVP~\cite{MVP} uses a CNN for amortized decoding of small primitives that can be compared to the Gaussians in 3DGS~\cite{3dgs}, but typically at lower sharpness. %
One approach to extend 3DGS to dynamic scenes is to optimize parameters like positions over time~\cite{luiten2023dynamic}, resulting in a representation that is hard to control. This limitation can be overcome by animating a canonical 3DGS representation with a deformation field~\cite{yang2023deformable3dgs}, similar to the Nerfies~\cite{park2021nerfies} approach. NPGA adopts this paradigm too.

\subsection{3D Morphable Models}

Traditional morphable models~\cite{blanz1999morphable, paysan20093d} learn a representation of body geometry via PCA. They are one of the primary tools to drive human animation and are heavily used in industry applications, making them a core building block for work on virtual avatars. While some models are dedicated to specific regions like the face~\cite{paysan20093d} and head~\cite{FLAME}, some variants include the neck~\cite{HACK} or even the entire body~\cite{SMPL-X:2019}. More recently, neural equivalents of 3DMMs, such as i3DMM~\cite{i3DMM}, ImFace~\cite{ImFace} or NPHM~\cite{nphm,mononphm}, have improved upon the expression fidelity compared to classical PCA-based models. They encode a canonical representation of the geometry via a signed distance functions (SDF), which can be mapped to arbitrary expressions via a deformation field. NPGA utilizes NPHM as it offers several beneficial characteristics: It models the face densely including eyes, hair, and teeth, it captures local details well and it disentangles shape from expressions. %

Instead of replacing mesh-based 3DMMs altogether, several approaches have been proposed to model details on top of an underlying PCA-driven mesh \cite{auth_ava,FaceScape,FaceVerse}.
Specifically, \citet{auth_ava} leverage high-frequency information from tracked UV texture maps to create and animate avatars, using a high-detailed, learned expression space.

\subsection{Human Head Reconstruction and Animation}

Existing approaches for animating avatars mainly differ in two fundamental aspects: first, the utilized 3D representation in combination with its rendering mechanism, and second, how the expression codes are transferred into scene dynamics.
Existing work has explored, among others, meshes in combination with deferred neural rendering \cite{nha,deep_video_portraits}, neural radiance fields \cite{gafni2021nerface, havatar, zielonka2022instant}, and CNNs in combination with primitive-based volume rendering \cite{neural_volumes,MVP,auth_ava}. 
Recently, there has been a lot of interest in point-based representations and rendering \cite{Zheng2023pointavatar,GA,GHA,dhamo2023headgasrealtimeanimatablehead}, especially since \citet{3dgs} proposed 3D Gaussian Splatting.%
While some approaches choose to explain the face's movement explicitly through the underlying mesh of a 3DMM, e.g., \cite{Athar_2022_CVPR_rignerf,zielonka2022instant, GA}, others choose the opposing extreme of a more data-driven approach by freely learning the face movement using a neural component, which is directly conditioned on the expression codes \cite{gafni2021nerface,GHA,MVP}. 
For NPGA we adopt the latter idea but replace the 3DMM with a neural parametric model, anticipating that the improved tracking quality translates to higher fidelity avatar reconstructions and animations.

\section{Preliminaries}
\subsection{3D Gaussian Splatting (3DGS)}
3DGS uses a point-based scene representation, where each point represents a Gaussian primitive that is described by a position $\mu$, rotation $\mathbf{q}$, scale $\mathbf{S}$, opacity $\alpha$ and spherical harmonics coefficients $\mathbf{SH}$.
In the following, we let the following notation
\begin{equation}
    \mathcal{A} = \left\{ \mathbf{\mu}, \mathbf{q}, \mathbf{S}, \mathbf{\alpha}, \mathbf{SH} \right\}, \qquad I = \textsc{3dGS}\left(\mathcal{A}, \pi_{K, E}\right)
    \label{eq:static_scene_rep}
\end{equation}
denote the set of attributes $\mathcal{A}$ composing the Gaussian point cloud, and its tile-based differentiable rasterization into an image $I$ under the camera projection described by intrinsic and extrinsic parameters $K$ and $E$ respectively.

\subsection{Neural Parametric Head Models}
3DMMs describe the geometry (and appearance) of faces (or heads) using disentangled parametric spaces for identity and expression variations. 
NPHM is a special case of a 3DMM, which represents a person's head geometry using a neural SDF and deformation field conditioned on identity latent codes $\zid$ and expression codes $\zexp$, respectively.
In particular, our work builds on top of MonoNPHM formulation, which describes expressions using a neural backward deformation field 
\begin{equation}
        x_c = \B(x_p; ~ \zid,  \zexp)
\end{equation}
 that warps points $x_p$ in posed space into canonical space $x_c$.

\section{Method}

\cref{fig:method_overview} shows an overview of our proposed representation and methodology to build our Neural Parametric Gaussian Avatars (NPGA). As described in \cref{sec:npga}, our avatars are composed of two key components: a canonical Gaussian point cloud $\mathcal{A}_c$ and a dynamics module $\mathcal{D}$ which deforms the Gaussians when provided with an expression code, similar to recent work ~\cite{GA,GHA,dhamo2023headgasrealtimeanimatablehead}.
In \cref{sec:cycle} we describe our distillation strategy that allows NPGA to leverage the rich latent expression space and detailed motion prior of MonoNPHM~\cite{mononphm}. 
Given multi-view video recordings alongside tracked MonoNPHM expression codes, we jointly optimize for our canonical Gaussians and dynamics module, as described in \cref{sec:optimization}.

\subsection{Neural Parametric Gaussian Avatars}
\label{sec:npga}
\subsubsection{Canonical Representation}
Compared to the default scene representation of 3DGS, outlined in \cref{eq:static_scene_rep}, we augment our canonical Gaussian point cloud 
\begin{equation}
    \mathcal{A}_c = \left\{ \mathbf{\mu}, \mathbf{q}, \mathbf{S}, \mathbf{\alpha}, \mathbf{SH},   \right\} \cup \left\{ \mathbf{f} \right\}
    \label{eq:can_rep}
\end{equation}
with per-Gaussian features $\mathbf{f} \in \mathbb{R}^{N \times 8}$. While these features are static themselves, they provide crucial semantic information to describe the dynamic behavior of the respective primitives. 
In some sense, our per-Gaussian features serve a similar purpose as positional encodings \cite{mildenhall2021nerf, mueller2022instant}, %
which are uncorrelated with spatial coordinates, have infinite spatial resolution and do not require additional data structures.
The idea of per-Gaussian features has been previously proposed in \cite{GHA,dhamo2023headgasrealtimeanimatablehead}. Crucially, however, both works do not utilize these features to better predict the movement of Gaussians, which we ablate in \cref{sec:ablation_study}. \cite{dhamo2023headgasrealtimeanimatablehead} goes beyond static per-primitive features by dynamically blending them with expression codes. In the formulation of \cite{GHA} and ours, this happens implicitly inside the MLP introduced in the next paragraph.

\subsubsection{Dynamics Module}

We model facial expressions using a dynamics model $\mathcal{D}$ which is decomposed into two Multi-Layer Perceptrons (MLPs), a coarse prior-based network $\F$ and a network $\G$ responsible for modeling all remaining details.
Our prior-guided forward deformation field
    \begin{equation}
        \delta_{\mu}^{\F} = \F(\mu, \mathbf{f}; ~\zexp) \in \mathbb{R}^3
        \label{eq:F}
    \end{equation}
is optimized to act as the inverse of MonoNPHM's backward deformations $\B$, as we describe later in
\cref{sec:cycle}. 
$\F$ is a coordinate-based network which predicts offsets $\delta_{\mu}^{\F}$ to the Gaussian centers $\mu$ and is conditioned on spatial coordinates $\mu$, features $\mathbf{f}$ and expression code $\zexp$.
Note, that $\F$ acts independently on each primitive, which we omit in \cref{eq:F} and below for clarity.

To represent dynamics beyond NPHM's prior, such as fine-scaled expression-dependent wrinkles, and appearance changes, e.g. due to ambient occlusions and changes in blood flow concentration, we rely on a second MLP
    \begin{equation}
        \delta_a^{\G} = \G_a(\mu, \mathbf{f}; ~ \zexp) \quad (\forall a \in \mathcal{A}_c),
    \end{equation}
which predicts offsets for all canonical Gaussian attributes $a$. We use the same architecture for $\G$ as for $\F$, besides having more output channels due to the increased number of attribute offsets.
In total, given an expression code $\zexp \in \mathbb{R}^{100}$ we obtain the Gaussian point cloud in posed space 
$\mathcal{A}_p$ by adding $\delta^{\F}$ and $\delta^{\G}$ to their respective canonical attributes, which we denote as
\begin{equation}
    \mathcal{A}_p = \mathcal{D}(\mathcal{A}_c;~ \zexp).
\end{equation}

\subsubsection{Screen-Space Refinement}

Finally, after rendering the posed Gaussians $\mathcal{A}_p$ using the differentiable rasterizer from \citet{kerbl3Dgaussians}, we apply a screen-space CNN network~\cite{latent_avatar}:
\begin{equation}
    [\hat{I}_{\text{rgb}}, I_{\text{h}}] = \textsc{3dGS}(\mathcal{A}_p; ~ \pi_{K, E}), \quad \hat{I}_{\text{cnn}} = \text{CNN}([\hat{I}_{\text{rgb}}, I_{\text{h}}]),
\end{equation}

where the $\hat{I}_{\text{rgb}}$ denotes rendered RGB colors.
$I_{h}$ is a latent image used for the CNN refinement module, which is obtained by rendering $\mathbf{h} + \delta_{\mathbf{h}}^{\G}$, where $\mathbf{h}$ are latent CNN features which we additionally include in $\mathcal{A}_c$.
We took inspiration from \citet{GHA}, to include a screen-space CNN, but note that we decided against performing super-resolution.

\subsection{Cycle-Consistency Distillation}
\label{sec:cycle}

One of our core ideas is to leverage MonoNPHM's motion prior and expression space. 
However, since it utilizes a \emph{backward} deformation field $\B$, which warps points into canonical space, we cannot directly incorporate this deformation prior in our pipeline. Instead, we need \emph{forward} deformations, which warp points into posed space, such that they can be directly rasterized.
While it is possible to numerically approximate the inverse of $\B$ using iterative root-finding~\cite{chen2021snarf,fastsnarf}, we search for a more computationally efficient method.
To this end we resort to a cycle consistency loss, similar to \cite{scanimate} and follow-ups \cite{dhamo2023headgasrealtimeanimatablehead,Yang_2022_CVPR_banmo} which optimize for volumetric skinning weights used for animation via linear blend skinning.

We propose to directly distill a forward deformation network $\F$ as the inverse of $\B$ (see \cref{fig:method_overview}, using a cycle consistency loss
\begin{equation}
    \mathcal{L}_{\text{cyc}}(x_c) = \left\Vert  \B\left( \F(x_c, \mathbf{f}(x_c))\right) - x_c \right\Vert_2^2.
    \label{eq:cyc}
\end{equation}
Using \cref{eq:cyc} we can directly supervise $\F$ with the knowledge of $\B$ for arbitrarily sampled points $x_c\in\mathbb{R}^3$ in canonical space and expression codes $\zexp$. Note, that in \cref{eq:cyc} we omit the dependence on expression codes $\zexp$ for clarity.
During this stage, the canonical space is not yet discretized into a set of Gaussian primitives. Hence, we utilize a feature \emph{field}
\begin{equation}
    \mathbf{f}(x_c) = \textsc{TriPlane}(x_c)
\end{equation}
represented as low-resolution 64x64 triplanes~\cite{conv_occ_nets,Chan2022_eg3d}, which can be evaluated at arbitrary points $x_c$ in canonical space.
For convenience, we train $\F$ once, using all six identities from the NeRSemble dataset~\cite{nersemble} that we perform experiments with. 

During training, each person has their own $\textsc{TriPlane}$, which we regularize using a total variation loss. Furthermore, we regularize the norm of predicted offsets $\Vert \F(x_c, \mathbf{f}(x_c)\Vert_2$ to be small.
Note that distilling $\F$ separately results in insignificant performance changes.

\subsection{Avatar Optimization Strategy}
\label{sec:optimization}
After obtaining a forward deformation field $\F$ using our cycle-consistency distillation strategy, we aim to jointly optimize for the canonical parameters $\mathcal{A}_c$ and MLP $\G$ to minimize a photometric energy term.
To initialize the canonical Gaussians centers $\mu$ we sample $50.000$ points uniformly on the iso-surface of the tracked MonoNPHM model. The per Gaussian features are initialized by querying $\textsc{TriPlane}(\mu)$ at the sampled Gaussian centers. All remaining attributes are initialized using the default 3DGS procedure.
In practice, we observed that keeping $\F$ frozen results in sub-optimal performance, which is likely caused through topological issues during distillation in the mouth region.
Hence, we decide to further optimize $\F$ alongside $\G$ and $\mathcal{A}_c$, however, using a significantly smaller learning rate and a warm-up schedule that encourages the preservation of the distilled prior.

Our optimization strives to minimize the photometric data term
\begin{equation}
    \mathcal{L} = \Vert I - \hat{I}_{\text{rgb}}\Vert_1 + \lambda\left(1\!-\!{\footnotesize \text{SSIM}}(I, \hat{I}_{\text{rgb}})\right) + \lambda\left(1\!-\!{\footnotesize \text{SSIM}}(I, \hat{I}_{\text{cnn}})\right),
\label{eq:loss}
\end{equation}
where $I$ denotes a randomly sample ground truth image from the NeRSemble dataset with corresponding expression codes $\zexp$.

\subsubsection{Regularization}

We find that regularizing both our canonical representation $\mathcal{A}_c$, as well as our dynamics module $\mathcal{D}$ is crucial to avoid overfitting to the training expressions.
To regularize NPGA we utilize a Laplacian smoothness term based on the $k$NN graph of canonical Gaussian centers. 
To this end let
\begin{equation}
    \mathcal{R}_{\text{lap}}(x) = \left\Vert \frac{1}{|\mathcal{N}_i|} \left( \sum_{j \in \mathcal{N}_j}x_j\right) - x_i\right\Vert_2^2 \quad \left( 
    x \in \{\textbf{f}\} \cup \bigcup_{a\in\mathcal{A}_c}\delta_a^{\G}
    \right),
\end{equation}

denote a Laplacian smoothness term, which we use to regularize the per Gaussian features $\mathbf{f}$, as well as, the offset predictions $\delta_a^{\G}$ for all attributes $a\in \mathcal{A}_c$.
Note, that whenever the number of canonical Gaussians changes due to densification or pruning, we recompute the $k$NN graph.

In addition to these smoothness terms, we  encourage $\F$ and $\G$ to predict small offsets
\begin{equation}
    \mathcal{R}_{\delta} ~=~ \lambda_{\mu}^{\F}\Vert \delta_{\mu}^{\F}\Vert_2^2 ~~ + 
        \sum_{a \in \mathcal{A}}\lambda_a^{\G} \Vert  \delta_a^{\G} - \mathbf{e}_a \Vert_2^2,
\end{equation}
where $\mathbf{e}_a$ denotes the neutral element for the group operation acting on attribute $a$.
Similarly, we impose regularization on the per Gaussian attributes $\mathcal{R}_{\mathbf{f}} = \Vert \mathbf{f} \Vert_2^2$ to remain small, and utilize the scale regularization loss of \cite{saito2024rgca}, which punishes scales $S$ lying outside of a well-behaved range.

\newcolumntype{Y}{>{\centering\arraybackslash}X}
\newcolumntype{P}[1]{>{\centering\arraybackslash}p{#1}}

\begin{figure*}%
    \centering
    \includegraphics[width=0.92\textwidth]{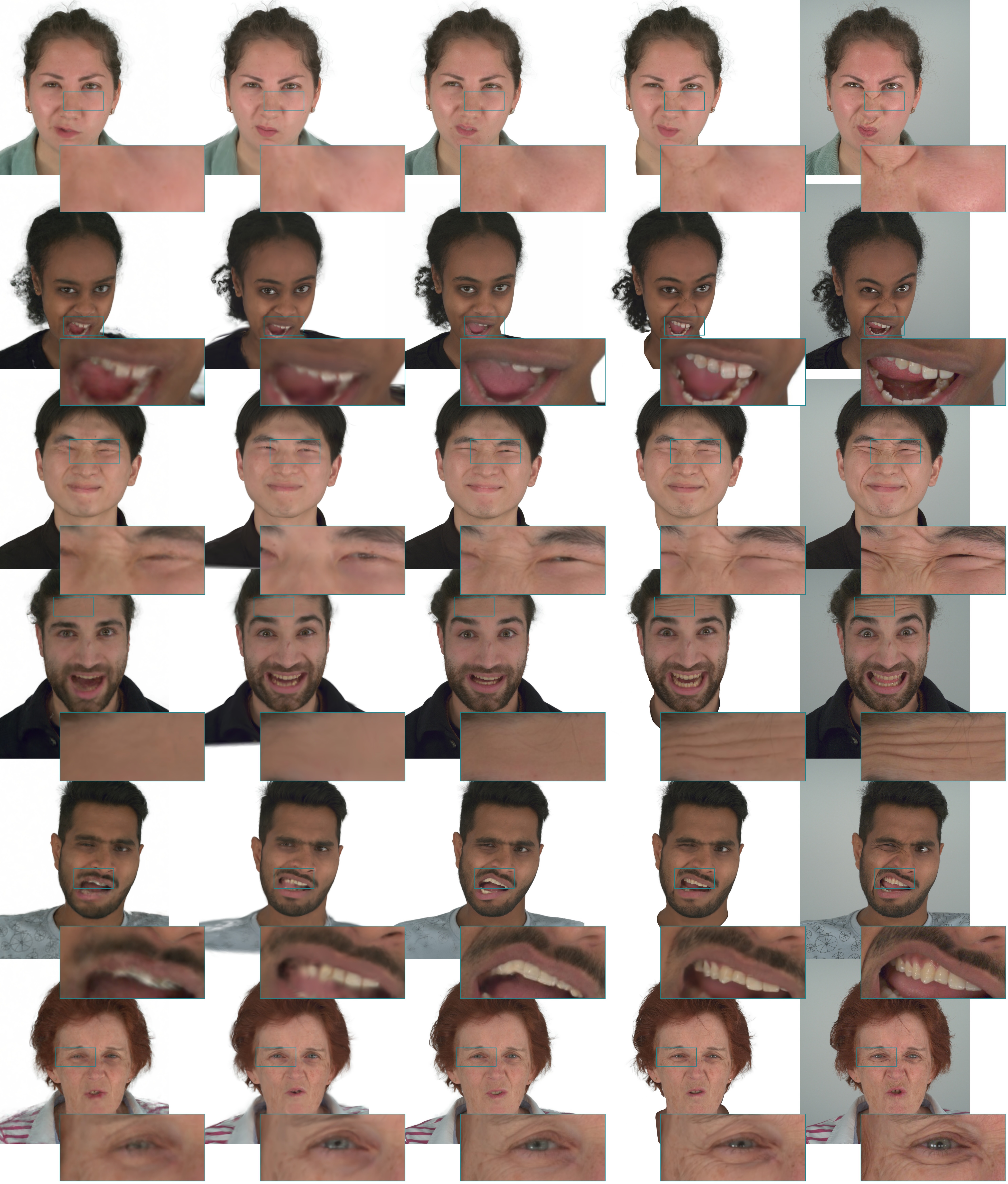}

     \begin{tabularx}{\textwidth}{
         P{0.22\textwidth}
         P{0.17\textwidth}
         P{0.15\textwidth}
         P{0.15\textwidth}
         P{0.15\textwidth}
     }
    \small{MVP} & 
    \small{GaussianAvatars} &
    \small{GHA} & 
    \small{Ours} & 
    \small{Ground Truth}
    \end{tabularx}
    \caption{
    \textbf{Self-Reenactment:} Qualitative comparison of different methods on the held-out sequence.
    }
    \label{fig:self_reenactment}
\end{figure*}

\subsubsection{Adaptive Density Control (ADC)}

A central ingredient to the success of 3DGS is its strategy to adaptively add and prune Gaussians in areas where they are needed or redundant, based on a set of simple yet effective heuristics that are periodically invoked.
The rules of ADC have been designed with static scenes in mind and we find the default settings to be suboptimal for our avatar creation. 
In the dynamic scenario, there can be areas that remain hidden for large parts of the training sequence, such as the mouth interior. Therefore, we adjust the ADC by employing a generalized mean
\begin{equation}
    M_i^{e} = \left( \frac{1}{N}\sum_{t\in T} \tau_{\text{t}}^e\right)^{1/e}
    \label{eq:generalized_mean}
\end{equation}
to aggregate the view-space gradients $\tau_{t}$ of the $i$-th primitive of all frames $T$ between invocations of the ADC mechanism. Note, that $e=1$ results in the default 3DGS settings. By increasing the exponent $e$ the aggregation $M_i^e$ becomes closer to a maximum function. Therefore, a few visible frames can be sufficient for the ADC to trigger densification, which is especially important for regions like the teeth and mouth interior.
We find that $e=2$ already results in an increased number of Gaussians, leading to more detailed reconstruction, especially in the mouth interior.

Furthermore, we replace the hard opacity reset mechanism of 3DGS, which we find to be harmful to our optimization, with a softer variant proposed in \cite{bulo2024revising}.
Instead of infrequently setting the $\alpha$ values to be almost transparent, the opacities get reduced frequently for a small amount only, i.e. in our experiments by $0.01$.

\subsection{Differences to GaussianHeadAvatar (GHA)}
\label{sec:diff_gha}
In general our avatar formulation shares large parts with that of GHA~\cite{GHA}.
A major difference is our use of MonoNPHM as underlying 3DMM, which we enabled through our distillation strategy. Furthermore, we utilize the per-Gaussian features to also influence the prediction of $\delta_{\mu}$, we add a Laplacian regularization, do not rely on screen-space super resolution or perceptual losses, and modify the ADC strategy instead of resorting to a fixed number of Gaussians as \citet{GHA}.

\section{Results}

For our experiments, we use the state-of-the-art, public multi-view video NeRSemble dataset~\cite{nersemble}, from which we choose a diverse set of six subjects performing challenging facial expressions.
After providing additional details on the performed experiments and baseline methods in \cref{sec:baselines,sec:implementation_details}, we present our main results on the tasks of self- and cross-reenactment in \cref{sec:self_reenactment,sec:cross_reenactment}, respectively.
Finally, in \cref{sec:ablation_study} we ablate a series of experiments validating our proposed model components.
We highly encourage the reader to consult our supplemental video for complete results including temporal information.

\subsection{Experimental Setup}
\label{sec:baselines}

The NeRSemble dataset provides 16 synchronized and calibrated videos, from which we choose 15 cameras for training and the frontal camera for evaluation.
Furthermore, we train our avatars on all sequences, except for the "FREE"-sequence which we keep as a held-out evaluation sequence for the self-reenactment task.

\subsubsection{Baselines}
\paragraph{GaussianAvatars (GA)~\cite{GA}:~ } 
GaussianAvatars is a recent method that creates 3DGS-based avatars, by binding them to the FLAME 3DMM model~\cite{FLAME}. Therefore, GaussianAvatars can be extremely efficiently animated, since there is no need to evaluate a costly neural component. On the downside, GaussianAvatars are limited to the facial movements lying inside the FLAME expression space.
\paragraph{GaussianHeadAvatar (GHA) \cite{GHA}:~ }
GHA is another recent 3DGS-based avatar method, which also learns deformation fields from multi-view video and is controlled through their custom multi-view BFM~\cite{paysan20093d} tracking. 
In general, our approach is similar to GHA. Thus we list the major differences in \cref{sec:diff_gha}.
Furthermore, we add a version of GHA which is conditioned on our tracked MonoNPHM expression codes, indicated as GHA$_{\text{NPHM}}$.
\paragraph{Mixture of Volumetric Primitives (MVP) \cite{MVP}:~}
MVP utilizes a combination of volume rendering and a head-geometry aware CNN that creates a volumetric payload in an amortized fashion. In contrast to our method and the other baselines, MVP utilizes a Variational Auto-Encoder (VAE)~\cite{Kingma2014_vae} to learn a latent expression encoding based on their 3DMM tracking.
Note, however, that we do not provide MVP with view-average textures to obtain a more comparable evaluation setting.
\begin{table}[htb]
\caption{
    \textbf{Quantitative Comparison:} We compare against our baselines on self-reenactment using a held-out sequence. For completeness we also report metrics on the held-out camera of the training sequences, denoted as novel-view synthesis (NVS).
    }
    \label{tab:main_results}
\begin{tabular}{@{}lcccccc@{}}
\toprule
\multirow{2}{*}{Method} & \multicolumn{3}{c}{NVS}              & \multicolumn{3}{c}{Self-Reenactment} \\ \cmidrule(l){2-7} 
                        & PSNR$\uparrow$ & SSIM$\uparrow$ & LPIPS$\downarrow$                  & PSNR$\uparrow$       & SSIM$\uparrow$       & LPIPS$\downarrow$      \\ \midrule
MVP                     & 33,42    & 0,957    & \multicolumn{1}{c|}{0,083} & 27.19          & 0.919          & 0.114          \\
GA                      & 32,95    & 0,956    & \multicolumn{1}{c|}{0,080} & 27.77          & 0.926          & 0.104          \\
GHA                     & 33,92    & 0,953    & \multicolumn{1}{c|}{0,045} & 26.81          & 0.914          & 0.077          \\

\textbf{Ours}           &\textbf{36,84}	    &\textbf{0,971}    & \multicolumn{1}{c|}{\textbf{0,034}} & \textbf{30.26}          & \textbf{0.934}          & \textbf{0.055}          \\ \hline

GHA$_{\text{NPHM}}$     & 33,09    & 0,952    & \multicolumn{1}{c|}{0,049} & 26.60          & 0.911          & 0.078          \\
Ours$_{\text{BFM}}$     &  34,97    &  0,962    & \multicolumn{1}{c|}{ 0,052} & 28.82          & 0.924          &  0.076    
\\ \bottomrule
\end{tabular}

\end{table}

\newcolumntype{Y}{>{\centering\arraybackslash}X}
\newcolumntype{P}[1]{>{\centering\arraybackslash}p{#1}}

\begin{figure*}
    \centering
    \includegraphics[width=\textwidth]{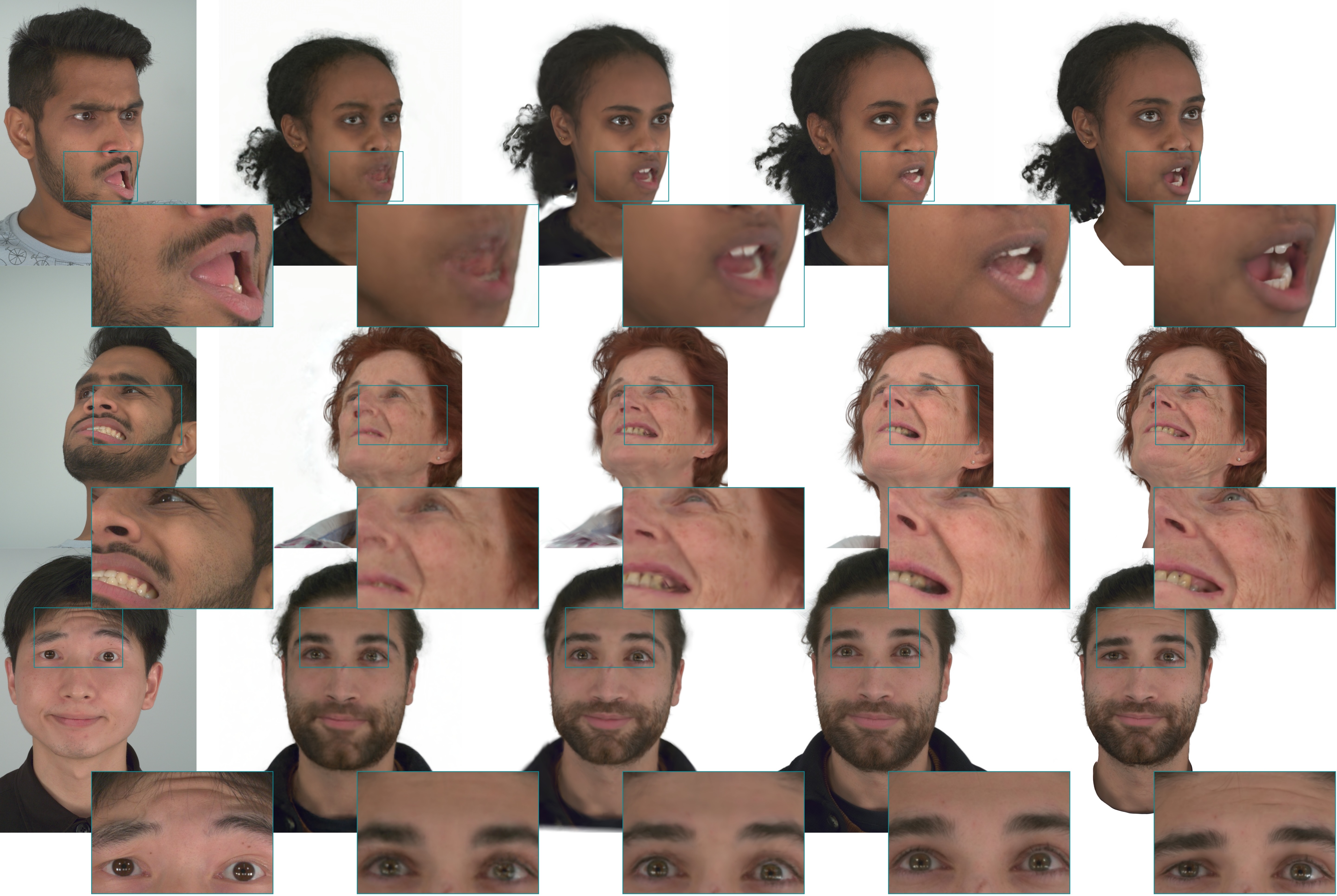}
     \begin{tabularx}{\textwidth}{
         P{0.15\textwidth}
         P{0.19\textwidth}
         P{0.20\textwidth}
         P{0.20\textwidth}
         P{0.20\textwidth}
     }
    \small{Driving Expression } &
    \small{MVP} & 
    \small{GaussianAvatars} &
    \small{GHA} & 
    \small{Ours}
    \end{tabularx}
    \caption{
    \textbf{Cross-Reenactment:} Qualitative comparison of transferring a driving expression from a different identity (left) to an avatar.
    }
    \label{fig:cross_reenactment}
\end{figure*}

\subsubsection{Metrics}

To evaluate the self-reenactment task we use the Peak Signal-to-Noise Ratio (PSNR), structural similarity index measure (SSIM), and perceptual LPIPS~\cite{lpips} metrics. For the sake of completeness, we also report numbers for a dynamic novel view synthesis (NVS) scenario, where we compare all methods on the held-out camera view of the training sequences.
We focus our evaluation on the facial region, since neck and torso are not accurately explained by NPHM and the underlying 3DMMs of our baselines. To this end, we leverage segmentation masks from Facer~\cite{zheng2022farl} to mask out the neck and torso before computing the metrics.
Furthermore, we compute metrics at a resolution of 550x802. Since we train GHA on 1024x1024 we downsample and crop the generated images accordingly.

\subsection{Implementation Details}
\label{sec:implementation_details}

\paragraph{Hyper-Parameters}
For both our deformation networks $\F$ and $\G$ we use 6-layer MLPs with a hidden dimensionality of $256$.
In order to preserve the prior that $\F$ obtained in our distillation procedure, we set its learning rate to $4e-5$, while $\G$ is equipped with a much higher learning rate of $2e-3$. Additionally, we freeze the network parameters of $\F$ for the first $5.000$ optimization steps.
We decay both learning rates twice by a factor of $2$ during the course of $800.000$ optimization steps.
Furthermore, we employ weight decay on $\F$ and $\G$, using a weight of $0.1$ as an additional regularization measure.
We perform an ADC step every $5.000$ iterations, and multiply the gradient threshold for the densification by a factor of $2$, to accommodate for the fact that our loss combines the losses of the RGB rendering and CNN-refined predictions.

\newcolumntype{Y}{>{\centering\arraybackslash}X}
\newcolumntype{P}[1]{>{\centering\arraybackslash}p{#1}}
\begin{figure*}[htb]

    \centering
    \includegraphics[width=\textwidth]{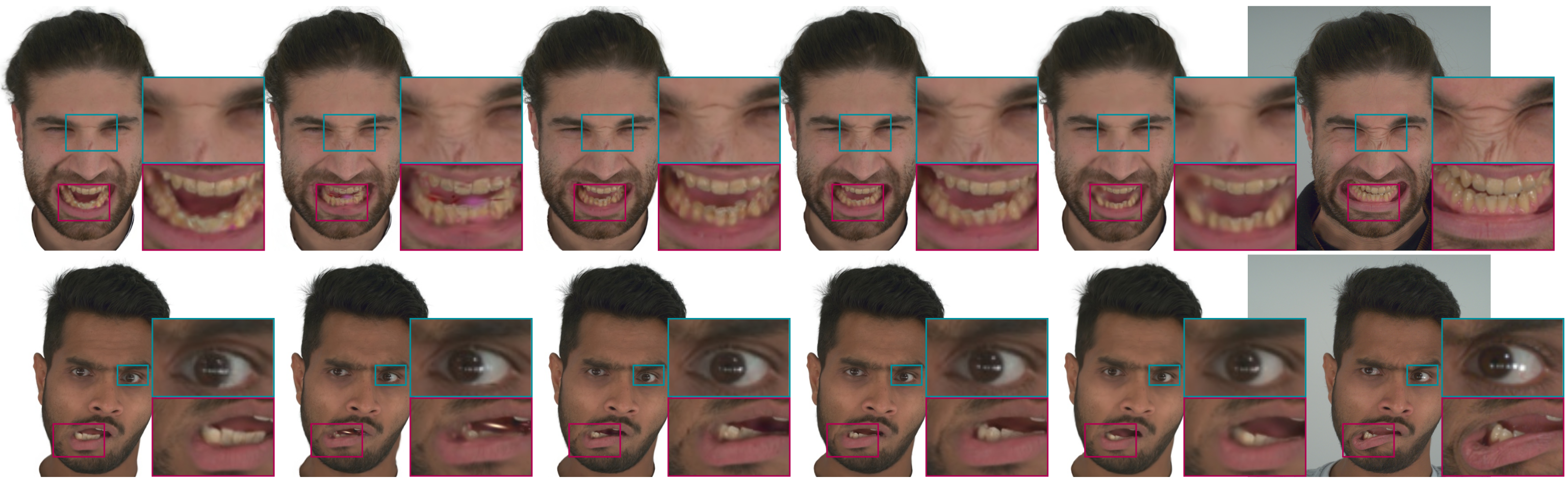}

     \begin{tabularx}{\textwidth}{
         P{0.16\textwidth}
         P{0.15\textwidth}
         P{0.14\textwidth}
         P{0.16\textwidth}
         P{0.10\textwidth}
         P{0.12\textwidth}
     }
    \small{Vanilla} & 
    \small{ + p.G.F} &
    \small{ + Lap. smoothness} & 
    \small{Ours} & 
    \small{Ours - ADC} & 

    \small{Ground Truth }
    \end{tabularx}
    \caption{
    \textbf{Ablation Study:}
    Without utilizing per Gaussians features ("Vanilla"), the avatars fail to represent fine expression details and complicated regions like the eyes and bottom teeth. Adding per Gaussian features (p.G.F.) results in significantly sharper reconstructions but is prone to artifacts under extreme expressions. Adding our Laplacian regularization ("+Lap. smoothness") and a screen-space CNN ("Ours") finally resolves all artifacts.
    Furthermore, "Ours-ADC" demonstrates that the default densification strategy inhibits detailed reconstructions.
    }
    \label{fig:ablation}

\end{figure*}

\paragraph{Runtime}
While we do not focus on efficient training, animation, or rendering of avatars, we acknowledge the importance of fast animation and rendering. In our unoptimized implementation, we can render images at 31 frames per second~(FPS) for 550x802 and 18~FPS at~1100x1604 on an NVIDIA RTX3080 graphics card, which includes deformation, rendering, and CNN. When omitting the CNN the speed increases to 43 and 38~FPS, respectively. As a comparison, GHA runs at 22~FPS at 1024x1024 on the same machine.
\cref{fig:num_gaussians} indicates the number of Gaussians during training, which we limit to a maximum of 250k.
We train all our avatars, and baselines, until convergence, which roughly takes 7~hours for GA (on an RTX2080), 30~hours (on an RTX3090) for GHA and our method, and 60~hours (on an RTX2080) for MVP.

\paragraph{Data Preparation}
We obtain MonoNPHM trackings on the NeRSemble dataset using a purely geometric constraint between the MonoNPHM's predicted surface and a point cloud reconstructed using COLMAP~\cite{schoenberger2016sfm}. We utilize the same tracking algorithm that has been previously used in \cite{diffusionavatars,FaceTalk}. For training and quantitative evaluation, we use a resolution of 550x802, the same as we use for MVP and GA. For our qualitative results, we fine-tune our avatars on 1100x1604 resolution for another 5~hours of training time.
Furthermore, we mask out the torso, since it is neither contained in NPHM's expression space nor the focus of our work.

\subsection{Self-Reenactment}
\label{sec:self_reenactment}
\begin{figure}
    \vspace{-0.1cm}
    \centering
    \includegraphics[width=\columnwidth]{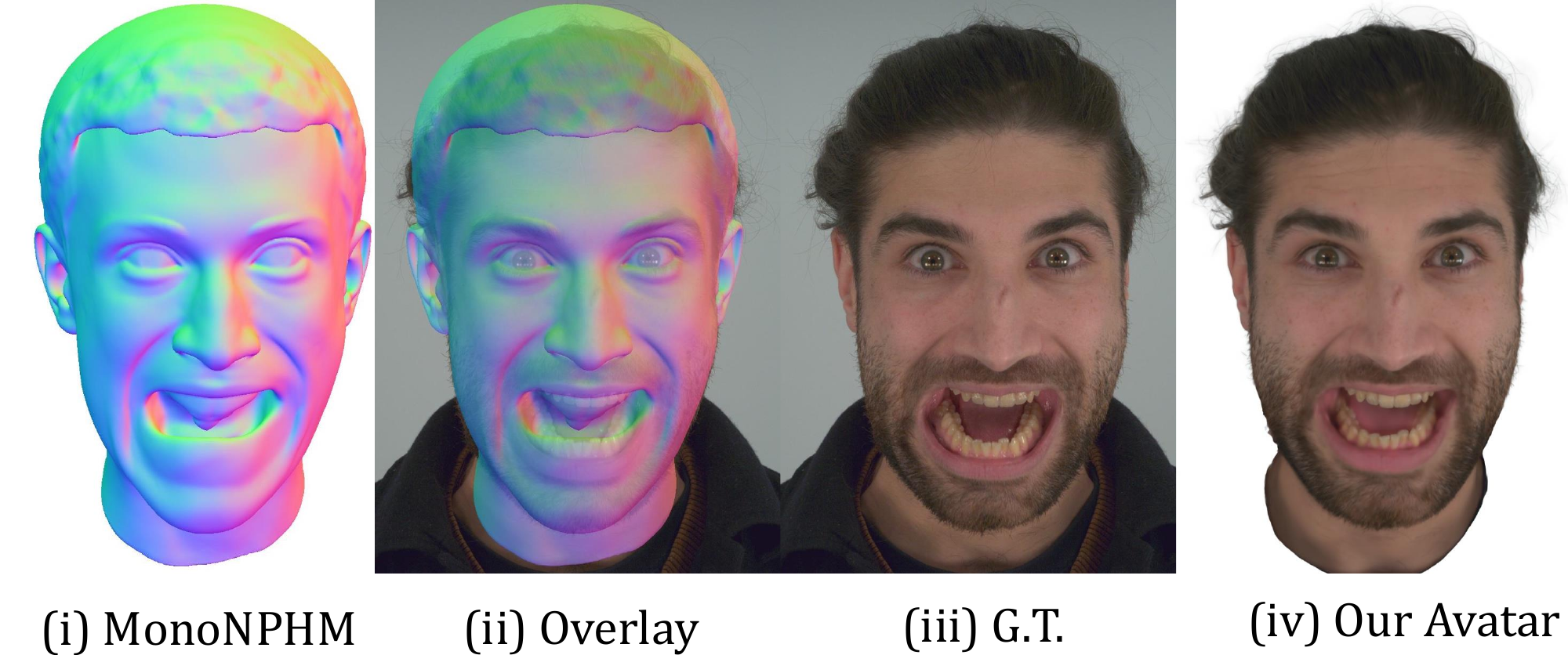}
    \caption{\textbf{Tracking Failure}: While our MonoNPHM-based tracking occasionally fails for extreme facial expressions (see normals (i) and overlay (ii)), NPGA still produces good animations for held-out expressions (see (iv)) due to the smooth nature of expression latent space.
    }
    \label{fig:tracking_failure}
        \vspace{-0.2cm}

\end{figure}

Our main evaluation is concerned with the self-reenactment task.
For this purpose, all avatars are trained on a set of 21 training sequences alongside their respective tracking results. 
To evaluate the avatars, they are animated using the tracked expressions from a held-out test sequence.
We present qualitative and quantitative results in \cref{fig:self_reenactment} and \cref{tab:main_results}, respectively, and recommend the reader to consider the supplemental video for temporal results. 
Our predicted self-reenactments portray the unseen expression more accurately and contain sharper details in relatively static areas like the hair region.
\paragraph{Importance of the Underlying Tracking}
Interestingly, GHA$_{\text{NPHM}}$ performs slightly worse than GHA, indicating that MonoNPHM expression codes alone do not immediately boost performance. Instead, we hypothesize that without NPHM's motion prior as initialization, NPHM's latent expression distribution might provide a more complicated training signal compared to the linear blendshapes of BFM.
Additionally, we train our avatars, but utilize the BFM tracking from GHA, denoted as Ours$_{BFM}$. Indicating that our overall framework still slightly outperforms GHA when using the same tracked expression codes.
In \cref{fig:tracking_failure} we show that the learned dynamics can still result in accurate self-reenactments, despite flaws in the tracking as long as there is sufficient correlation the associated latent codes.

\subsection{Cross-Reenactment}
\label{sec:cross_reenactment}
\begin{figure}[htb]
    \centering
    \includegraphics[width=\columnwidth]{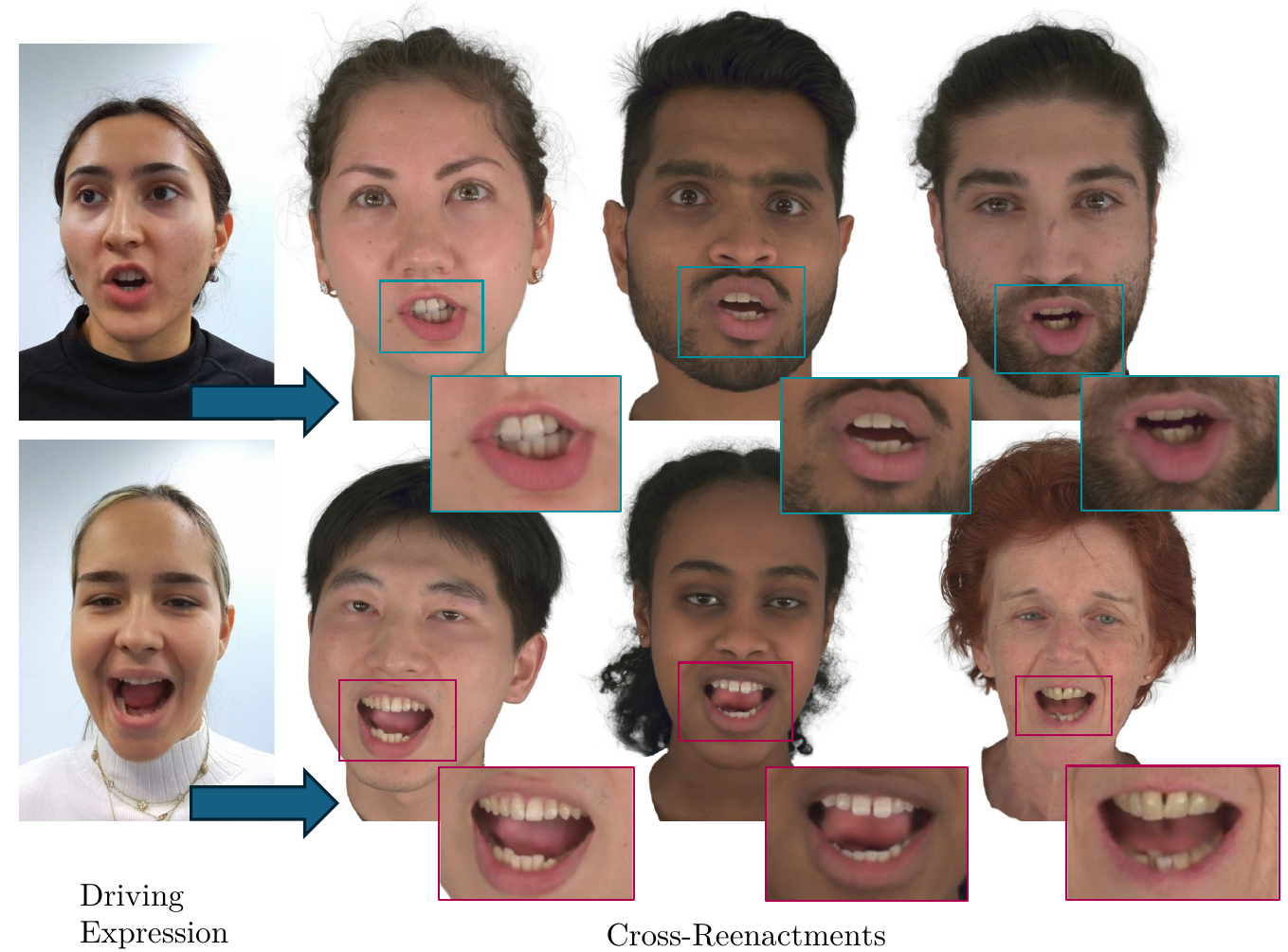}
    \caption{\textbf{Real-World Application}: We utilize the monocular RGB tracking from MonoNPHM to animate our high-fidelity avatars, demonstrating the applicability of our avatars outside of multi-view capture studios.
    }
    \label{fig:kinect}
\end{figure}

Another crucial task is cross-reenactment, where driving expressions from another person are transferred to the avatar. Since a ground truth for cross-reenactment does not exist, we only report a qualitative comparison, which is presented in \cref{fig:cross_reenactment}. 
We observe that all methods successfully disentangle identity and expression information, allowing for effective cross-reenactment. Our avatars, however, preserve the most details from the driving expressions.
To demonstrate real-world applicability, \cref{fig:kinect} depicts cross-reenactment animations of our avatars using monocular RGB videos from a commodity camera under real-world circumstances. 
To this end, we utilize the monocular MonoNPHM tracker proposed by \citet{mononphm}.
\subsection{Ablations Study}
\label{sec:ablation_study}
In order to verify several important components of \OURS, we perform ablation experiments using three subjects.
Quantitative and qualitative results of our ablations can be found in \cref{tab:ablation_results} and \cref{fig:ablation}, respectively.
First, we run a "vanilla" version of NPGA that serves as a baseline. This version does not utilize per Gaussian features, the CNN, Laplacian smoothness terms, and uses $e=1$ in \cref{eq:generalized_mean} for the ADC. This model fails to produce sharp renderings for fine-scale details, and areas that are complicated due to frequent occlusions and reflections, like the bottom teeth and eyes.
\begin{table}[]
\caption{
    \textbf{Ablations:} We perform our ablation experiments on a subset of three subjects. We report Novel-View Synthesis (NVS) for completeness.
    }
    \label{tab:ablation_results}
\begin{tabular}{@{}lcccccc@{}}
\toprule
\multirow{2}{*}{Method} & \multicolumn{3}{c}{NVS}              & \multicolumn{3}{c}{Self-Reenactment} \\ \cmidrule(l){2-7} 
                        & PSNR$\uparrow$ & SSIM$\uparrow$ & LPIPS$\downarrow$                  & PSNR$\uparrow$       & SSIM$\uparrow$       & LPIPS$\downarrow$      \\ \midrule
Vanilla                 & 35.80    & 0.965    & \multicolumn{1}{c|}{0.048} & 30.16	&0.927	&0.067          \\
+PGF                   & 37.04    & 0.970    & \multicolumn{1}{c|}{0.037} & 30.54	&0.929	&0.059          \\
+Lap.smooth             & 36.85    & 0.969    & \multicolumn{1}{c|}{0.038} & 30.56	&0.928	&0.059         \\
\hline
\textbf{Ours(+CNN)}                    & \textbf{37.23}    & \textbf{0.972}    & \multicolumn{1}{c|}{\textbf{0.033}} & \textbf{30.65}	&\textbf{0.933}	&\textbf{0.053}          \\
\hline
Ours-ADC              & 36.12    & 0.967    & \multicolumn{1}{c|}{0.045} & 30,49  &0,933	&0.070         \\
Ours-$\delta_{\mu}$              & 36.73    & 0.968    & \multicolumn{1}{c|}{0.040} & 30,44  &0,931	&0.059         \\

 \bottomrule
\end{tabular}

\end{table}

\begin{figure}
    \vspace{-0.2cm}

     \centering
     \includegraphics[width=\columnwidth]{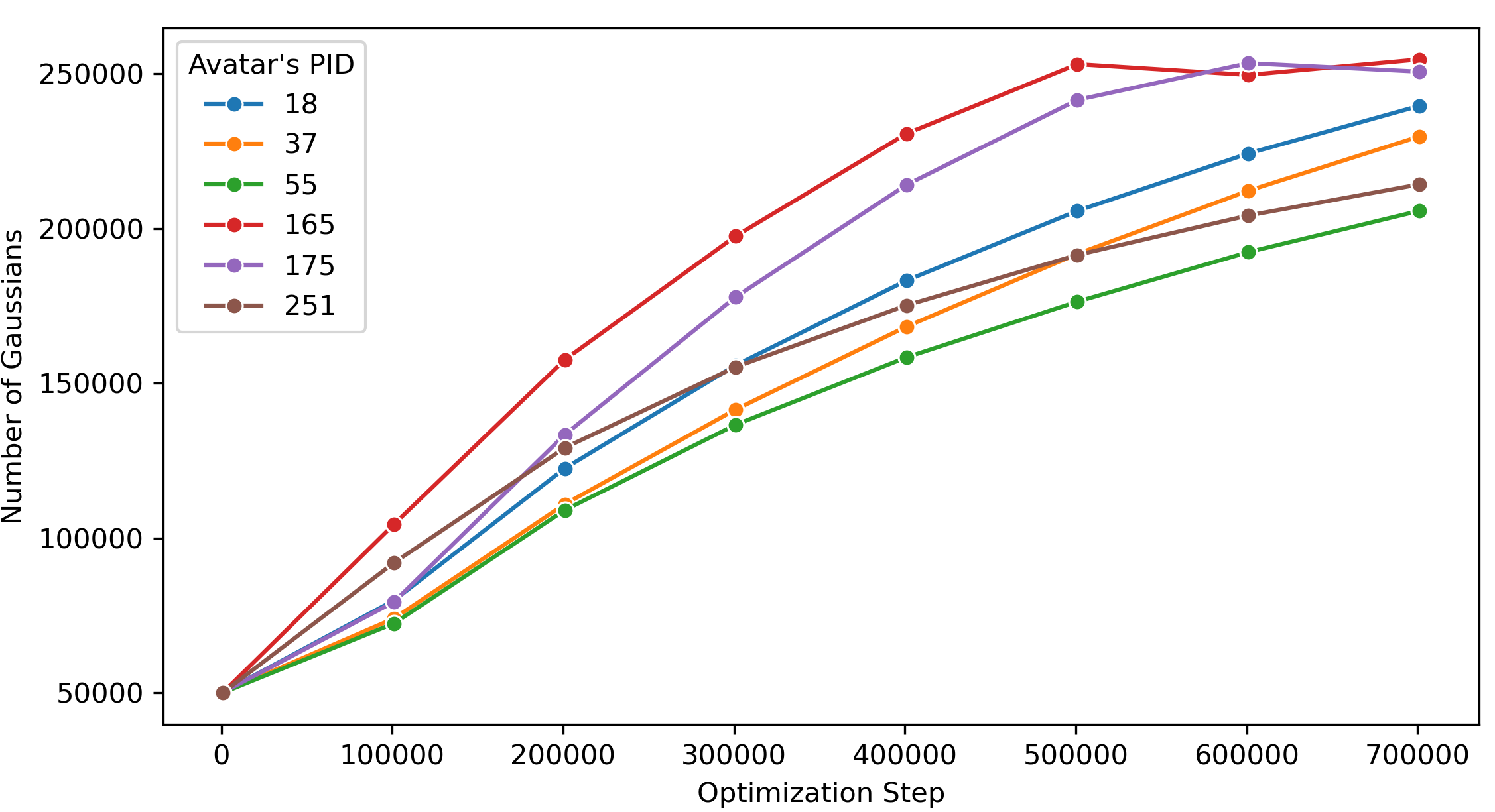}
    
    \caption{\textbf{Number of Gaussians:}
    We illustrate the growing number of Gaussians for each of our avatars, where PID denotes the personal identifiers in the Nersemble dataset. We limit the maximum number to 250k.
    }
    \label{fig:num_gaussians}
        \vspace{-0.2cm}

\end{figure}

\paragraph{Per-Gaussian Features}
When adding per-Gaussian features to the vanilla model, denoted as {"+p.G.F."}, the increased representational capacity results in sharper reconstructions. At the same time, we occasionally observe artifacts of "free-floating" primitives, as highlighted in the second column of \cref{fig:ablation}.
These artifacts can be largely removed using our proposed Laplacian smoothness terms. \cref{tab:ablation_results} indicates that using this regularization significantly shrinks the generalization gap between training (NVS) and testing (self-reenactment) sequences.
Compared to this model, "Ours" also includes a CNN, which further boosts metrics and visual quality.
Additionally, we ablate the importance of letting per-Gaussian features influence the predicting of position offsets, since recent work \cite{GHA,dhamo2023headgasrealtimeanimatablehead} does not allow such interactions. In \cref{tab:ablation_results} {"Ours-$\delta_{\mu}$"} denotes a version of {\OURS} that disables such an influence, indicating the benefits of our formulation.
\paragraph{Adaptive Density Control}
Finally, we show the importance of using an adjusted ADC strategy. 
Surprisingly, using the default ADC settings with a densification interval of 5000 steps and an opacity reset interval of $50.000$ steps almost completely diminishes the improvements of our other contributions.
While we do not claim that using $e=2$ in \cref{eq:generalized_mean} is necessary for great performance, we simply note that finding a setting that lets enough Gaussians appear is important, especially for fine-scaled wrinkles and teeth. 
The progress of the number of Gaussian during the optimization is illustrated in \cref{fig:num_gaussians}.


\section{Limitations and Future Work}

In our experiments, we show that {\OURS} can create controllable and high-fidelity virtual head avatars from multi-view video data. 
However, both controllability and reconstruction quality of our avatars are fundamentally restricted to what the underlying 3DMM can explain.
Therefore, regions like the neck, torso, tongue, and eyeball rotation, which are not explained by NPHM's expression codes, cannot be animated as reliably or might even lead to artifacts due to overfitting. 
Possible solutions are extensions of the underlying 3DMM to provide a more complete description of a person's state, e.g. the inclusion of the neck~\cite{HACK} or even torso and complete bodies~\cite{SMPL-X:2019}.

Furthermore, as a data-driven approach to avatar creation, our method is limited, to some degree, to the available training data per person. We believe that recent large-scale multi-view video dataset of human heads~\cite{nersemble,pan2024renderme} open up opportunities to learn a generalized head model, such as \cite{auth_ava}, with much higher fidelity than NPHM and other available 3DMMs, through the use of photometric optimization and efficient rendering, like 3DGS.

\section{Ethical Considerations}
The creation of photo-realistic avatars has potential for a wide range of malevolent activities, e.g. privacy violation, identity theft and spreading of deceptive content through deepfakes.
We condemn any malicious or unauthorized uses of such technology, and emphasize the need for reliable detection methods that maintain the authenticity of media content and advert negative societal impact.

\section{Conclusion}

In this work, we have proposed Neural Parametric Gaussian Avatars (NPGA), a method for creating accurately controllable and high-fidelity virtual head avatars.
The main focus of our work is the usage of MonoNPHM's rich expression space and motion prior. To this end, we leverage a cycle-consistency strategy to distill a forward deformation field from MonoNPHM, such that it becomes compatible with 3D Gaussian Splatting.
We proposed an effective regularization strategy and adjust the adaptive density control strategy, both of which are important for ideal avatar quality.
We show that utilizing per-Gaussian features to conditioning the complete dynamics module helps, compared to limiting the influence to appearance changes.
In our experiments, we significantly outperform the previous state-of-the-art avatars on self-reenactment. 
Finally, we showed the applicability of our avatars beyond a controlled multi-view set-up by animating them from monocular RGB video trackings.
\subsubsection*{Acknowledgements}
This work was funded by Synthesia and supported by the ERC Starting Grant Scan2CAD (804724), the German Research Foundation (DFG) Research Unit ``Learning and Simulation in Visual Computing''. We would like to thank our research assistant Mohak Mansharamani, and Angela Dai for the video voice-over.

\bibliographystyle{ACM-Reference-Format}
\bibliography{sample-bibliography}

\end{document}